\documentclass[letterpaper, 10pt, conference]{ieeeconf}
\IEEEoverridecommandlockouts

\usepackage{amsmath,amssymb,amsfonts}
\usepackage{algorithm,algorithmic}
\usepackage{graphicx}
\usepackage{textcomp}
\usepackage{xcolor}
\usepackage{cite}
\usepackage{url}

\usepackage{multirow,multicol}
\usepackage{makecell}
\usepackage{threeparttable}
\usepackage{soul}
\usepackage{colortbl}
\usepackage{balance}

\title{\LARGE \bf
SVIA: A Street View Image Anonymization Framework for Self-Driving Applications
}

\author{Dongyu Liu$^{1,*}$, Xuhong Wang$^{2,*}$, Cen Chen$^{1,\dag}$, Yanhao Wang$^{1}$, Shengyue Yao$^{2}$,  Yilun Lin$^{2}$%
\thanks{$^{*}$Both authors contributed equally to this work.}%
\thanks{$^{\dag}$Corresponding author of this work.}%
\thanks{$^{1}$D. Liu, C. Chen, and Y. Wang are with the School of Data Science and Engineering, East China Normal University, Shanghai, China.}%
\thanks{$^{2}$X. Wang, S. Yao, and Y. Lin are with Shanghai Artificial Intelligence Laboratory, Shanghai, China.}%
}

\begin{document}

\maketitle

\begin{abstract}
In recent years, there has been an increasing interest in image anonymization, particularly focusing on the de-identification of faces and individuals.
However, for self-driving applications, merely de-identifying faces and individuals might not provide sufficient privacy protection since street views like vehicles and buildings can still disclose locations, trajectories, and other sensitive information.
Therefore, it remains crucial to extend anonymization techniques to street view images to fully preserve the privacy of users, pedestrians, and vehicles.
In this paper, we propose a \underline{S}treet \underline{V}iew \underline{I}mage \underline{A}nonymization (SVIA) framework for self-driving applications.
The SVIA framework consists of three integral components: a \emph{semantic segmenter} to segment an input image into functional regions, an \emph{inpainter} to generate alternatives to privacy-sensitive regions, and a \emph{harmonizer} to seamlessly stitch modified regions to guarantee visual coherence.
Compared to existing methods, SVIA achieves a much better trade-off between image generation quality and privacy protection, as evidenced by experimental results for five common metrics on two widely used public datasets.
\end{abstract}

\section{Introduction}

The proliferation of Artificial Intelligence (AI) techniques has raised the compelling need to address concerns about \emph{data privacy} in their applications.
To mitigate such concerns, several privacy protection methods, e.g., federated learning \cite{federated}, differential privacy \cite{differential}, and anonymization \cite{dataanonymization}, have been proposed and widely applied.
Generally, these methods leverage user-generated data for AI applications while safeguarding the privacy of individuals.
However, most of them are specific to structured data and thus face unique challenges when applied to (unstructured) images and videos.

Self-driving is one of the application scenarios in which \emph{images} and \emph{videos} are massively and rapidly generated while \emph{privacy} has been a crucial issue.
Specifically, a self-driving vehicle can generate approximately 1.4 TB of data per hour \cite{flashmemory}, including a large number of street view images that might contain sensitive information.
For privacy concerns and regulations, it is important to anonymize street view images to prevent violations arising from the inclusion of sensitive information within them in the deployment of autonomous driving systems.
Here, sensitive information refers to those that can potentially identify individuals, vehicles, locations, and trajectories, e.g., human faces and bodies, license plates, buildings, and street signs.
Anonymized street view images can be shared and published with little privacy concern \cite{deid}, thus providing high-utility images in real scenarios to train machine learning models for self-driving applications.

Unfortunately, there is a significant gap in the current body of research on image anonymization, as most of them mainly focus on face de-identification \cite{anonymousnet, deepprivacy, ciagan}.
This gap becomes particularly apparent in the context of autonomous driving, where effective anonymization is required for all forms of sensitive information, including but not limited to faces and human bodies.
Although some recent studies use street view image datasets for experimentation \cite{ldfa, deepprivacy2}, they also employ anonymization only on humans.
It is still possible to extract sensitive information from the supposedly ``anonymized'' street view images generated by these methods.
Consequently, there is still a risk of leaking locations, trajectories, personal properties, and other personally identifiable information.
Traditional de-identification methods, such as adding mosaics to obscure all sensitive information, often result in serious image degradation and are unsuitable for autonomous driving scenarios since they can lead to much inferior model performance and cause safety issues.

Currently, most image anonymization methods rely on generative adversarial networks \cite{gan} (GANs) because they better preserve the validity of images than traditional methods while effectively erasing sensitive information.
However, GANs still exhibit several limitations, such as slow convergence, mode collapse, and the need for well-designed generators and discriminators, when used for image anonymization.
These issues prevent them from generating high-resolution images and often compromise image diversity, rendering them less suitable for street-view images with complex semantic information.
In addition, GANs also lack controllability and struggle to generate images under specified conditions.
Very recently, a few attempts have been made to employ diffusion models \cite{diffusion} for image anonymization because they are shown to be more effective than GANs in terms of scalability and capability to generate images under given conditions \cite{dms_beat_gans}.
Therefore, in this paper, we opt to extend diffusion models for street view image anonymization.

\vspace{1mm}
\noindent\textbf{Our Results.}
In this paper, we propose a novel \underline{S}treet \underline{V}iew \underline{I}mage \underline{A}nonymization (SVIA) framework.
The SVIA framework consists of three integral components: a \emph{semantic segmenter} to segment a street view image into functional regions, an \emph{inpainter} based on the latent diffusion model \cite{stablediffusion} (LDM) to generate alternatives to privacy-sensitive regions, and a \emph{harmonizer} to seamlessly stitch modified regions to ensure visual coherence.
Our SVIA framework effectively addresses the challenges encountered by conventional non-ML approaches, including the need for precise control over image semantics, the generation of high-quality output images, and the maintenance of consistency in generating meaningful entities throughout the anonymization process.
Moreover, to overcome the limitations associated with existing methods based on GANs and to effectively tackle the concerns specific to street view image anonymization, we leverage LDMs, which involves utilizing text and image prompts as conditional controls to generate sanitized street view images, enabling the anonymization of diverse semantic categories.
Our contributions are summarized as follows.
\begin{itemize}
    \item We propose a novel three-component street view image anonymization (SVIA) framework.
    Unlike existing approaches that focus on face anonymization, SVIA concentrates on street view images and effectively addresses privacy concerns in the context of data collection for self-driving applications.
    \item We conduct extensive experiments on two widely used datasets: \emph{Cityscapes} and \emph{Mapillary Vistas}. The results demonstrate that SVIA better anonymizes both individuals and vehicles than several baselines. In particular, SVIA ensures location anonymization, making it difficult to infer the presence of a person or vehicle on a particular street or city based on the anonymized images. Moreover, SVIA still preserves the validity of the generated images, ensuring that the performance of downstream tasks is not compromised.
\end{itemize}

\section{Related Work}

\subsection{Image Anonymization}

Image anonymization aims to safeguard privacy by erasing sensitive information from images before publication.
Traditional techniques such as pixelation, blurring, and masking have been commonly used for this purpose, albeit at the cost of compromising image quality.
With the rapid development of deep learning techniques, a variety of methods based on deep neural networks have emerged for image anonymization, especially for face de-identification, which can be broadly classified into three categories.
The first category is related to \emph{facial attributes}.
Li and Lin \cite{anonymousnet} defined facial attributes as sensitive information and proposed a StarGAN model for anonymization.
Mirjalili et al.~\cite{privacynet} proposed a conditional GAN model to generate an anonymized facial image with specified attributes.
The second category is based on \emph{image inpainting}.
Hukkel\r{a}s et al.~\cite{deepprivacy} regarded face anonymization as a combination of facial detection and image inpainting.
They also provided an image inpainting method \cite{deepprivacy2} based on StyleGAN2 \cite{stylegan2} for full-body anonymization.
Klemp et al.~\cite{ldfa} considered using LDMs \cite{stablediffusion} to inpaint facial images with high facial detection rates and Fr\'{e}chet inception distance (FID) scores.
The third category pertains to \emph{face swapping}.
Gafni et al.~\cite{gafni} proposed an adversarial autoencoder-based method \cite{advae} for face swapping.
Maximov et al.~\cite{ciagan} used a conditional GAN with identity vectors as input for the identity transformation.
Ma et al.~\cite{cfanet} designed a GAN with an encoder-decoder generator to extract content and identity vectors and replace the latent identity vector to generate anonymized images.

We note that the aforementioned anonymization methods primarily target facial images and may not be suitable for other scenarios, such as street-view images. In addition, GAN-based methods require a significant amount of training data and computational resources, which is often very inefficient in practice.

\subsection{Semantic Segmentation}

For self-driving applications, the semantic segmentation of street view images is crucial to ensure the vehicle's proper functioning and accurate reception of external information.
Semantic segmentation is thus an important pre-processing step to facilitate image anonymization and other downstream tasks in self-driving applications.
Deep learning-based segmentation methods typically employ a UNet model \cite{unet} to extract image features, transform them into latent vectors, and decode them to generate pixel-wise masks for each semantic category.
DeepLabv3 \cite{deeplabv3} is an improvement over UNet for semantic segmentation.
Recently, Wang et al.~\cite{internimage} proposed InternImage based on the architecture of deformable convolution networks v3 (DCNv3).
This further produced better performance in semantic segmentation than UNet and DeepLabv3, while also exhibiting high accuracy in image classification and object detection.
Although semantic segmentation methods are not directly applicable to street view image anonymization, they can serve as a component in our SVIA framework to offer semantic information, which can guide and regulate other components to produce precise anonymization results.

\subsection{Diffusion-based Image Generation}

Image generation aims to create artificial images that do not exist in the real world.
It has recently been a popular topic in computer vision and graphics.
In addition to classic models based on autoencoders and GANs, diffusion models have emerged as the state of the art for image generation.
Denoising diffusion probabilistic models \cite{diffusion} and latent diffusion models \cite{stablediffusion} (LDMs) are representatives of diffusion models that are effective in generating high-quality images with intricate details and realistic textures.
Diffusion models allow for multi-modal conditions and various types of input such as images, texts, and audio to control the generation process.
Since diffusion models offer advantages in terms of scalability, versatility, and extensibility over other generative models, e.g., autoencoders and GANs, they will be used as the backbone component in our SVIA framework.

\subsection{Image Inpainting}

Image inpainting can be seen as a special case of image generation.
It takes the original image before restoration and the corresponding mask as input and generates a new image by modifying the regions covered by the mask in the original image as output, which has been used to remove occlusions, unwanted content, and sensitive information from images.
Zhao et al.~\cite{comodgan} proposed a co-modulation GAN model to balance the equality and diversity of image inpainting.
Li et al.~\cite{mat} proposed a mask-aware Transformer model that performed well in image inpainting.
LDM \cite{stablediffusion} provided better quality, flexibility, and controllability for image inpainting by allowing the generation of masked regions to be guided by textual prompts and has been widely used in image inpainting applications.
In this work, we adopted LDM-based image inpainting to generate anonymized images by replacing sensitive information with fake data, while preserving the integrity of non-sensitive data.

\section{Our Framework}

\begin{figure}[t]
    \centering
    \includegraphics[width=0.75\linewidth]{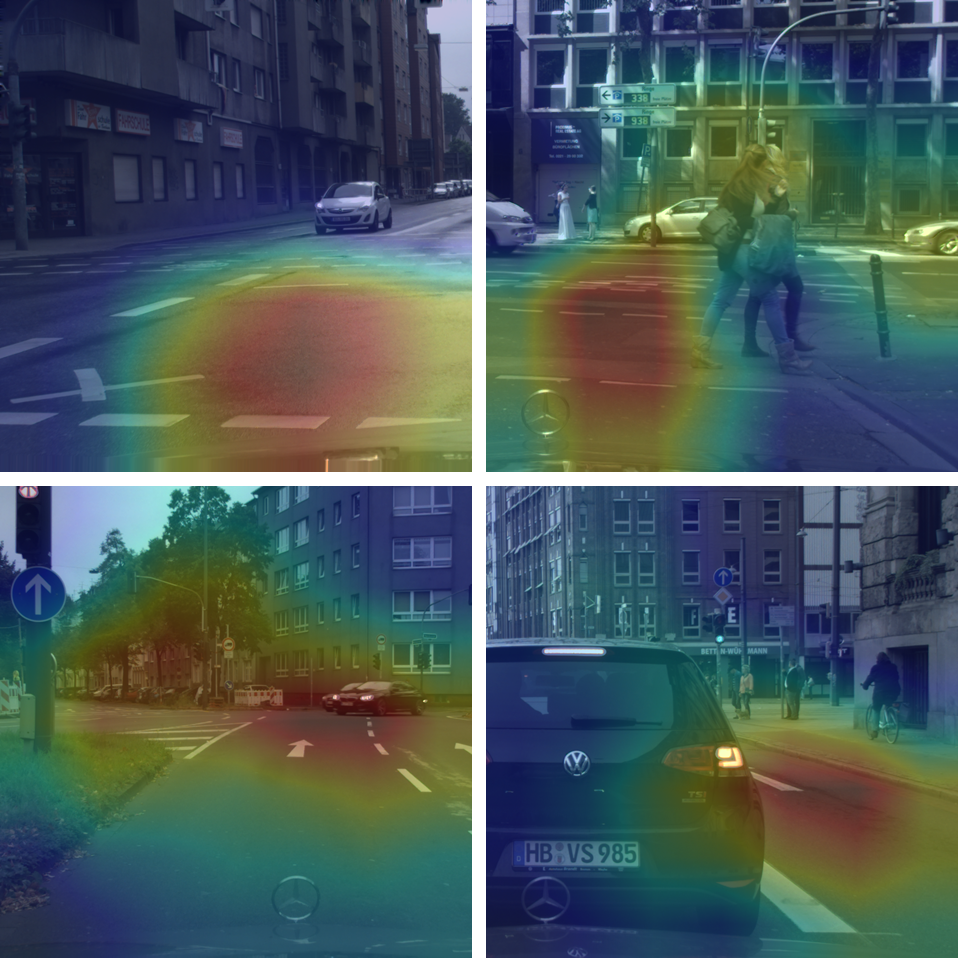}
    \caption{Illustration of Grad-CAM results for city classification, where the regions with higher scores for city re-identification are indicated in red and yellow colors. This reveals that roads and buildings are the primary semantic categories to identify the city where a street view image is taken.}
    \label{fig1}
\end{figure}

\begin{figure*}[t]
    \centering
    \includegraphics[width=0.98\linewidth]{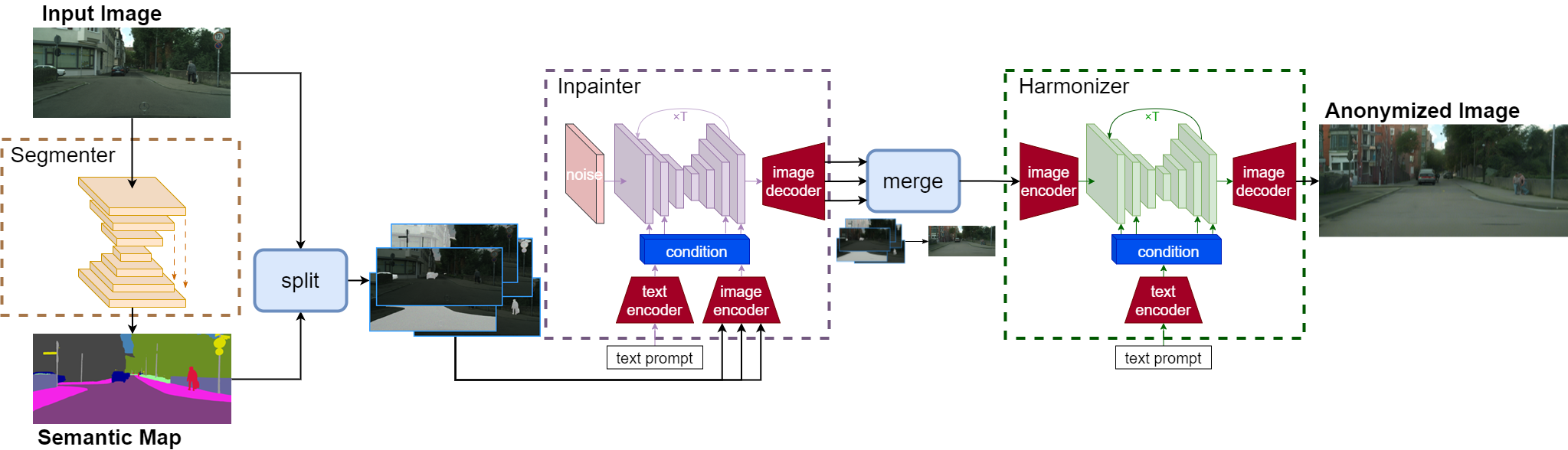}
    \caption{Overview of the SVIA pipeline.}
    \label{fig2}
\end{figure*}

\subsection{Background on Street View Image Anonymization}
\label{subsec-background}

There can exist multiple semantic categories within a street view image, such as \emph{sky}, \emph{plants}, \emph{vehicle}, and \emph{humans}.
We consider the following two aspects to remove any form of sensitive information from street view images.
\begin{itemize}
    \item \textbf{Semantic categories that cause \textit{direct} privacy leakage:} persons (e.g., walkers and bikers), vehicles (e.g., cars, trucks, and buses), and traffic signs can be used to identify a specific individual and track the trajectory. Therefore, it is crucial to anonymize these objects.
    \item \textbf{Semantic categories that can potentially cause \textit{indirect} privacy leakage:} Irregularities on roads, painted marks, and distinctive landmarks or architecturally significant buildings can reveal the cities of origin where images are taken. To verify this, we train a city classifier on the Cityscapes dataset with an accuracy of 99.2\%. Then, we utilize the Grad-CAM \cite{gradcam} method to identify the regions in an image that serve as distinguishing criteria for the re-identification of trajectories. The illustrative results in Fig.~\ref{fig1} align with our intuition.
\end{itemize}
Consequently, we choose the following five semantic categories in street view images to inpaint: \emph{person}, \emph{vehicle}, \emph{traffic sign}, \emph{road}, and \emph{building}.

\subsection{Overview of SVIA Framework}

We consider that street view images are captured by the cameras equipped with intelligent vehicles.
These images are valuable for the platform to train and evaluate self-driving models but also contain sensitive user information.
The platform needs high-utility images from users while protecting their privacy.
Therefore, before uploading the captured street view images to the server, our SVIA framework is activated to anonymize them.

Generally, the SVIA framework comprises three components: a \emph{segmenter}, an \emph{inpainter}, and a \emph{harmonizer}.
First, the segmenter is employed to semantically segment the input image.
The resulting segmented image consists of solid colors, where each color denotes a specific semantic category.
The masks w.r.t.~the semantic categories of \emph{person}, \emph{vehicle}, \emph{traffic sign}, \emph{road}, and \emph{building} are extracted from the segmented image.
To prevent privacy leakage, Laplacian noise is added to the original image in each mask \cite{ciagan}.
Second, noisy images with their corresponding masks are individually used as input for inpainting.
The inpainting process is performed separately on each of the five labeled images to remove sensitive information and generate plausible fake information.
The five fake images generated are then combined to create an inpainted image.
Third, the inpainted image is further processed by a shallow image-to-image harmonizer to eliminate any remaining sensitive information and repair errors caused by the combining process.
Finally, the generated images, which do not contain any sensitive information, are safe to be published and uploaded to the platform's server for downstream tasks with little privacy concern.
The above procedure is presented in Algorithm~\ref{alg-framework}.
Fig.~\ref{fig2} illustrates the overall SVIA pipeline.

\begin{algorithm}[t]
    \footnotesize
    \caption{SVIA}
    \label{alg-framework}
    \begin{algorithmic}[1]
        \renewcommand{\algorithmicrequire}{\textbf{Input:}}
        \renewcommand{\algorithmicensure}{\textbf{Output:}}
        \REQUIRE Image $x \in [0, 1]^{3 \times h \times w}$; number of semantic categories $n$; segmentation model $\Phi(x): [0, 1]^{3 \times h \times w} \rightarrow \{0, 1\}^{n \times h \times w}$; inpainting model $\Omega(\phi, x, pr): \{0, 1\}^{h \times w} \times [0, 1]^{3 \times h \times w} \times String \rightarrow [0, 1]^{3 \times h \times w}$; harmonizer $\Theta(x, pr): [0, 1]^{3 \times h \times w} \times String \rightarrow [0, 1]^{3 \times h \times w}$; text prompts for inpainting $prompt_i$ and harmonizer $prompt_h$.
        \ENSURE Anonymized image $y \in [0, 1]^{3 \times h \times w}$.
        \STATE $(\phi_1, \phi_2, \dots, \phi_n) \gets \Phi(x)$
        \FOR{$i = 1$ \textbf{to} $n$}
            \STATE $\tilde{x}_i \gets x + \phi_i\cdot\mathrm{Laplace}(0, 0.25)$
            \STATE $\hat{y}_i \gets \Omega(\phi_i, \tilde{x}_i, prompt_i)$
        \ENDFOR
        \STATE $\hat{y} \gets \sum_{i = 1}^{n}{(\phi_i \hat{y}_i)} + x \prod_{i = 1}^{n}{(1 - \phi_i)}$
        \STATE $y \gets \Theta(\hat{y}, prompt_h)$
        \RETURN $y$
    \end{algorithmic}
\end{algorithm}

\subsection{Semantic Segmentation}
\begin{figure*}[t!]
    \centering
    \includegraphics[width=0.98\linewidth]{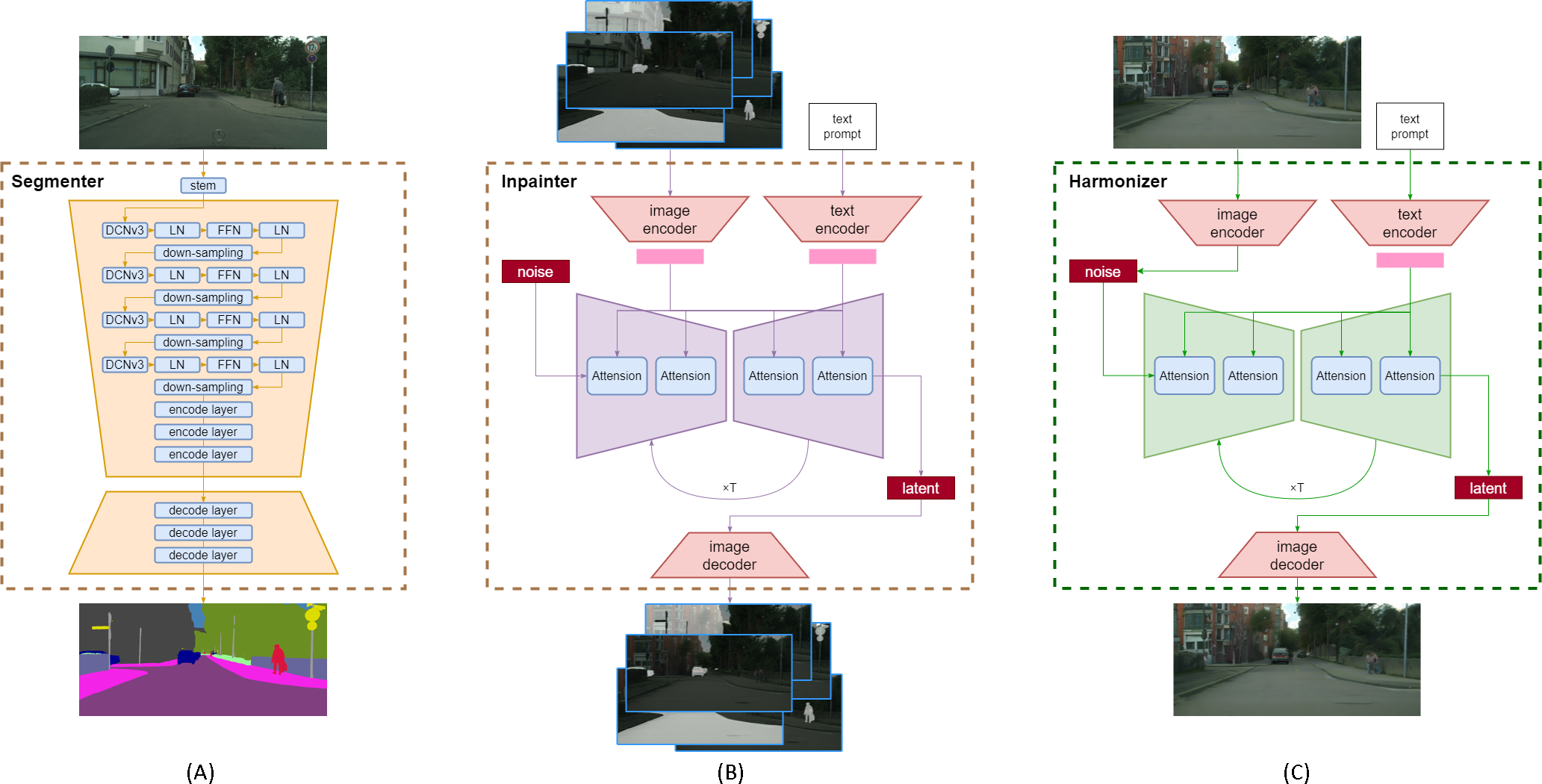}
    \caption{Detailed Architectures of (A) the segmenter, (B) the inpainter, and (C) the harmonizer.}
    \label{fig3}
\end{figure*}

Before anonymization is performed, it is necessary to obtain pixel-level masks that indicate the areas to be replaced.
This can be done by semantic segmentation, which involves the use of convolutional neural networks (CNNs) to output pixel-level multi-classification results.
In our case, a street view image should be segmented to create masks for persons, vehicles, traffic signs, roads, buildings, and other semantic categories.
These masks are helpful for downstream tasks, such as self-driving.
Inspired by the InternImage model \cite{internimage}, we utilize a segmenter based on the DCNv3 architecture, which includes a multi-head self-attention mechanism to enhance the model's ability to handle large-scale dependencies and adaptive spatial aggregation.
The semantic segmentation model extracts pixel-level masks for specific semantic categories of interest in the image.
These masks are then used as masking input for inpainting to cover privacy information.
The semantic segmentation model can be utilized to simulate downstream tasks and measure the impact of anonymization on image quality.
Fig.~\ref{fig3} (A) shows the architecture of the semantic segmentation model.

\begin{algorithm}[t!]
    \footnotesize
    \caption{Inpainting Model $\Omega(\phi, x, prompt)$}
    \label{alg-inpainting}
    \begin{algorithmic}[1]
        \renewcommand{\algorithmicrequire}{\textbf{Input:}}
        \renewcommand{\algorithmicensure}{\textbf{Output:}}
        \REQUIRE Masked noisy image $\tilde{x} \in [0, 1]^{3\times h\times w}$; mask $\phi \in \{0, 1\}^{n \times h \times w}$; text prompt $prompt$; text encoder $E_t : String \rightarrow \mathbb{R}^{s}$; image encoder $E_{img} : [0, 1]^{3 \times h \times w} \rightarrow \mathbb{R}^{s}$ and decoder $D_{img} : \mathbb{R}^{s} \rightarrow [0, 1]^{3 \times h \times w}$, denoising encoder $E_s(t) : \mathbb{R} \rightarrow \mathbb{R}^{s}$, UNet noise sampler $S(x, \dots) : \mathbb{R}^{s} \times \dots \rightarrow \mathbb{R}^{s}$; number of denoising steps $d$; latent size $s$; noise strength $\alpha_{1,\dots,d} \in (0, 1)$; constants in denoising steps $\sigma_{1, \dots, d}$.
        \ENSURE Inpainted image $y \in [0, 1]^{3 \times h \times w}$.
        \STATE $e_t \gets E_t(prompt)$
        \STATE $e_{img} \gets E_{img}(\tilde{x})$
        \STATE Sample noise $y_d \sim {N(0, 1)}^{s}$
        \FOR{$i = d$ \textbf{to} $1$}
            \STATE $e_s \gets E_s(i)$
            \STATE $\epsilon_i \gets S(y_i, e_s, e_t, e_{img})$
            \IF{$i \neq 1$}
                \STATE $z \sim {N(0, 1)}^{s}$
            \ELSE
                \STATE $z \gets \mathbf{0}^{s}$
            \ENDIF
            \STATE $\hat{y}_0^{(i)} \gets \frac{1}{\sqrt{\alpha_i}}(y_i - \sqrt{1 - \alpha_i}\epsilon)$
            \STATE $y_{i-1} \gets \sqrt{\alpha_{i - 1}}\hat{y}_0^{(i)} + \sqrt{1 - \alpha_{i-1} - \sigma_i^2} \epsilon_i + \sigma_i z$
        \ENDFOR
        \RETURN $y \gets D_{img}(y_0)$
    \end{algorithmic}
\end{algorithm}

\subsection{Image Inpainting}

To remove sensitive information from images, we employ image inpainting.
This involves masking the regions of the image that contain sensitive information and using the inpainting model to generate synthetic images that cover the corresponding regions in the original image.
The goal of inpainting is to erase sensitive information while preserving the utility and integrity of the image.
However, traditional image inpainting methods such as MAT\cite{mat} often produce low-quality images and compromise the validity of the image to some extent.
To address this issue, we propose a method that utilizes the Stable Diffusion v2 inpainting model \cite{stablediffusion}.
In our method, the input image is encoded and combined with the mask as a conditional input.
To integrate multi-modal features, we employ the attention mechanism.
A random noise vector is generated by an encoder and denoised by a noise sampler using the multi-modal conditional input.
Finally, the latent vector is decoded and output as the resultant inpainted image.
The detailed procedure is presented as Algorithm~\ref{alg-inpainting}.
Fig.~\ref{fig3} (B) describes the architecture of the image inpainting model.

\begin{algorithm}[t]
    \footnotesize
    \caption{Harmonizer $\Theta(x, prompt)$}
    \label{alg-harmonizer}
    \begin{algorithmic}
        \renewcommand{\algorithmicrequire}{\textbf{Input:}}
        \renewcommand{\algorithmicensure}{\textbf{Output:}}
        \REQUIRE Coarse image $x \in [0, 1]^{3 \times h \times w}$, text prompt $prompt$, text encoder $E_t : String \rightarrow \mathbb{R}^{s}$; image encoder $E_{img} : [0, 1]^{3 \times h \times w} \rightarrow \mathbb{R}^{s}$ and decoder $D_{img} : \mathbb{R}^{s} \rightarrow [0, 1]^{3 \times h \times w}$, denoising encoder $E_s(t) : \mathbb{R} \rightarrow \mathbb{R}^{s}$; UNet noise sampler $S(x, \dots) : \mathbb{R}^{s} \times \dots \rightarrow \mathbb{R}^{s}$; number of denoising steps $d$; latent size $s$; noise strength $\alpha_{1, \dots, d} \in (0, 1)$; constants in denoising steps $\sigma_{1, \dots, d}$.
        \ENSURE Harmonized image $y \in [0, 1]^{3 \times h \times w}$.
        \STATE $e_t \gets E_t(prompt)$
        \STATE $x_0 \gets E_{img}(x)$
        \STATE Sample noise $z \sim {N(0, 1)}^{s}$
        \STATE $y_d \gets \sqrt{\prod_{s=1}^{d} \alpha_s} x_0 + \sqrt{1 - \prod_{s=1}^{d} \alpha_s} z$
        \FOR{$i = d$ \textbf{to} $1$}
            \STATE $e_s \gets E_s(i)$
            \STATE $\epsilon_i \gets S(y_i, e_s, e_t, e_{img})$
            \IF{$i \neq 1$}
                \STATE $z \sim {N(0, 1)}^{s}$
            \ELSE
                \STATE $z \gets \mathbf{0}^{s}$
            \ENDIF
            \STATE $\hat{y}_0^{(i)} \gets \frac{1}{\sqrt{\alpha_i}}(y_i - \sqrt{1 - \alpha_i} \epsilon)$
            \STATE $y_{i-1} \gets \sqrt{\alpha_{i - 1}} \hat{y}_0^{(i)} + \sqrt{1 - \alpha_{i-1} - \sigma_i^2}\epsilon_i + \sigma_i z$
        \ENDFOR
        \RETURN $y \gets D_{img}(y_0)$
    \end{algorithmic}
\end{algorithm}

\subsection{Harmonizer}

Finally, we present \emph{harmonizer}, a component for generating harmonized images based on coarse images, which utilizes the stable diffusion model to blend and unify each image produced by the inpainting model.
The harmonizer is an image-to-image model designed to remove any noticeable ``hard lines'' resulting from the splicing of inpainted images and to further eliminate sensitive details.
Generally, it encodes the coarse image, introduces noise into the latent space before denoising it, and decodes the image into a finely generated one.
The procedure is presented as Algorithm~\ref{alg-harmonizer}.
Fig.~\ref{fig3} (C) depicts the architecture of the harmonizer.

\section{Experiments}

\subsection{Setup}

\noindent\textbf{Environment.}
All experiments were conducted on two Nvi-dia RTX 3090 GPUs with a batch size of 1 in each GPU.
Our code in the experiments, which has been open-sourced under the MIT license, can be accessed on GitHub through the link below:
\url{https://github.com/Viola-Siemens/General-Image-Anonymization}.

\vspace{1mm}
\noindent\textbf{Datasets.}
We used the following two datasets in the experiments:
(1) \emph{Cityscapes} \cite{cityscapes} is a collection of around 5,000 annotated street view images from 50 different cities and is used to evaluate the quality of image generation and privacy protection based on the city re-identification score;
(2) \emph{Mapillary Vistas} \cite{mapillary} contains 25,000 high-resolution street view images without location annotations and is also used to evaluate the quality of image generation and privacy protection based on image patch and person similarity scores.

\vspace{1mm}
\noindent\textbf{Baselines.}
We compared with three model-free anonymization methods: \emph{blurring}, \emph{pixelization}, and \emph{masking}.
We also compared with deep learning-based anonymization methods such as \emph{DeepPrivacy} \cite{deepprivacy}, and inpainting components such as \emph{MAT} \cite{mat} for ablation studies, which are both based on GANs.
We note that MAT is not suitable for anonymizing street view images, and we have made adaptations and incorporated the same harmonizer as SVIA to enhance the quality of images generated by MAT.

\subsection{Evaluation Metrics}
\label{subsec-metric}

A good image anonymization method should ensure (1) the high quality of image generation and (2) the outstanding ability of privacy protection.
The quality of image generation can be measured by the degree to which the accuracy of downstream tasks is affected by anonymization.
The ability of privacy protection is reflected by whether sensitive information is identifiable from the anonymized image.
Next, we describe the metrics used for evaluation in the two aspects.

\vspace{1mm}
\noindent\textbf{Quality of Image Generation.}
We use the following two metrics to evaluate the quality of generated images.
\begin{itemize}
    \item \textbf{Fr\'{e}chet Inception Distance (FID)} \cite{fid} is a metric for the difference between two images.
    Using an Inception v3 model, it extracts features from original and generated images and computes the difference in the mean and covariance of features.
    A lower FID score indicates higher quality and lower impacts on downstream tasks.
    \item \textbf{Kernel Inception Distance (KID)} \cite{kid} uses the maximum mean discrepancy (MMD) to measure the distinctions of the feature distributions between two images.
    KID better reflects some specific aspects of image quality, such as mode dropping or collapse.
    A lower KID score also indicates higher quality and lower impacts on downstream tasks.
\end{itemize}

\vspace{1mm}
\noindent\textbf{Ability of Privacy Protection.}
We use the following three metrics to evaluate the ability of each anonymization method for privacy protection.
\begin{itemize}
    \item \textbf{Accuracy of City Re-identification (ACR)} indicates how much location information is exposed in the generated image.
    We use the city classifier in Section~\ref{subsec-background} and Fig.~\ref{fig1} to classify the output images and calculate the classification accuracy as ACR.
    Lower ACR indicates a reduced possibility of trajectory disclosure and a better ability to protect location privacy.
    \item \textbf{Learned Perceptual Image Patch Similarity (LPIPS)} \cite{lpips} compares the $l_2$-distance of the feature maps of selected hidden layers between two images.
    A higher LPIPS score means that the features extracted from the original and generated images are less similar, thus implying a greater ability for privacy protection.
    \item \textbf{Person Similarity (PerSim)} measures if a person can be re-identified from the generated image. We adopt the widely used PASS method \cite{pass} to measure the similarity of persons. The lower the similarity, the more difficult it is to recognize the person in the generated image. Thus, a lower PerSim score indicates a stronger ability to protect the identity of individuals.
\end{itemize}

\subsection{Comparison of SVIA and Baselines}

\begin{table}[t]
    \centering
    \caption{Results for the generation quality and privacy protection ability of different methods}
    \label{tbl-1}
    \begin{center}
    \scalebox{0.72}{
    \begin{tabular}{c|c|c|c|c|c}
        \hline
        \multirow{2}*{Dataset} & \multirow{2}*{Method} & \multicolumn{2}{c|}{Image Quality} & \multicolumn{2}{c}{Privacy Protection}\\
        \cline{3-6}
        ~ & ~ & FID $\downarrow$ & KID $\downarrow$ & LPIPS $\uparrow$ & PerSim $\downarrow$\\
        \hline
        \multirow{7}*{Cityscapes} & w/o Anonymization & 0 & 0 & 0 & 1\\
        ~ & Pixelization & 63.37 (4) & 0.0641 (5) & 0.4306 (4) & 0.9121 (4)\\
        ~ & Blurring & 47.75 (3) & 0.0512 (3) & 0.3395 (5) & 0.9196 (5)\\
        ~ & GrayMask & 163.1 (6) & 0.1823 (6) & \textbf{0.5620 (1)} & 0.8845 (3)\\
        ~ & DeepPrivacy & \textbf{0.0273 (1)} & \textbf{0 (1)} & 0.0001 (6) & 0.9996 (6)\\
        ~ & MAT+Harmonizer & 67.46 (5) & 0.0628 (4) & 0.5286 (3) & \textbf{0.8532 (1)}\\
        ~ & SVIA (*Ours) & \cellcolor[rgb]{0.85, 0.85, 0.85}{37.74 (2)} & \cellcolor[rgb]{0.85, 0.85, 0.85}{0.0279 (2)} & \cellcolor[rgb]{0.85, 0.85, 0.85}{0.5320 (2)} & \cellcolor[rgb]{0.85, 0.85, 0.85}{0.8769 (2)}\\
        \hline
        \multirow{7}*{Mapillary Vistas} & w/o Anonymization & 0 & 0 & 0 & 1\\
        ~ & Pixelization & 86.23 (4) & 0.0443 (5) & 0.3129 (4) & 0.9028 (4)\\
        ~ & Blurring & 66.46 (3) & 0.0275 (3) & 0.2536 (5) & 0.9038 (5)\\
        ~ & GrayMask & 119.8 (6) & 0.0668 (6) & 0.3575 (3) & 0.8771 (3)\\
        ~ & DeepPrivacy & \textbf{0.0281 (1)} & \textbf{0 (1)} & 0.0001 (6) & 0.9681 (6)\\
        ~ & MAT+Harmonizer & 87.62 (5) & 0.0277 (4) & \textbf{0.4656 (1)} & \cellcolor[rgb]{0.85, 0.85, 0.85}{0.8630 (2)}\\
        ~ & SVIA (*Ours) & \cellcolor[rgb]{0.85, 0.85, 0.85}{65.01 (2)} & \cellcolor[rgb]{0.85, 0.85, 0.85}{0.0105 (2)} & \cellcolor[rgb]{0.85, 0.85, 0.85}{0.4556 (2)} & \textbf{0.8595 (1)}\\
        \hline
    \end{tabular}
    }
    \begin{tablenotes}
        \item[1] Note: The best and second-best values for each metric are highlighted in bold font and gray boxes, respectively; the numbers in parentheses indicate the ranking of an anonymization method for each metric.
    \end{tablenotes}
    \end{center}
\end{table}

\begin{table}[t]
    \centering
    \caption{Results for the accuracy of city re-identification on the Cityscapes dataset}
    \label{tbl-2}
    \begin{center}
    \scalebox{0.72}{
    \begin{tabular}{c|cccc}
        \hline
        Method & ACR@1 $\downarrow$ & ACR@2 $\downarrow$ & ACR@3 $\downarrow$ & ACR@4 $\downarrow$\\
        \hline
        w/o Anonymization & 99.2493\% & 99.6303\% & 99.7759\% & 99.8767\%\\
        Pixelization & 68.0878\% (4) & 85.6961\% (4) & 92.2241\% (4) & 95.0664\% (4)\\
        Blurring & 73.9110\% (5) & 89.5090\% (5) & 95.1242\% (5) & 97.4812\% (5)\\
        GrayMask & \textbf{18.8446\% (1)} & \textbf{27.4292\% (1)} & \textbf{33.9572\% (1)} & \textbf{40.2426\% (1)}\\
        DeepPrivacy & 99.2493\% (6) & 99.6303\% (6) & 99.7759\% (6) & 99.8767\% (6)\\
        MAT+Harmonizer & \cellcolor[rgb]{0.85, 0.85, 0.85}{29.2317\% (2)} & \cellcolor[rgb]{0.85, 0.85, 0.85}{43.5124\% (2)} & \cellcolor[rgb]{0.85, 0.85, 0.85}{53.4835\% (2)} & \cellcolor[rgb]{0.85, 0.85, 0.85}{60.7510\% (2)}\\
        SVIA (*Ours) & 29.5552\% (3) & 45.4535\% (3) & 56.4414\% (3) & 64.6678\% (3)\\
        \hline
    \end{tabular}
    }
    \begin{tablenotes}
        \item[1] Note: The meanings of bold font, gray boxes, and numbers in parentheses are the same as those of Table~\ref{tbl-1}.
    \end{tablenotes}
    \end{center}
\end{table}

We compare SVIA with the baselines on the Cityscapes and Mapillary Vistas datasets.
Table~\ref{tbl-1} presents the results for four quality and privacy metrics (except ACR) of different methods on the two datasets.
Table~\ref{tbl-2} reports the accuracy of city re-identification of each method on the Cityscapes dataset.
The results reveal that \emph{blurring} and \emph{pixelization} not only generate low-quality images but also provide poor privacy protection for street view images.
The \emph{gray masking} method effectively removes sensitive information, but significantly degrades image quality, as evidenced by its highest FID and KID scores.
All model-free methods fail to simultaneously achieve high image quality and good privacy protection.
DeepPrivacy produces high-quality images that are very similar to the original images, as indicated by the extremely low LPIPS scores.
This is further confirmed by high PerSim and ACR scores, suggesting that DeepPrivacy is not suitable for anonymizing street view images.
Compared to SVIA, the MAT inpainting component generates images of significantly lower quality, as signified by high FID and KID scores.
Meanwhile, its scores in terms of privacy protection are only marginally better than those of SVIA.
In summary, SVIA achieves the best trade-off between image generation quality and privacy protection among all compared methods.

\begin{figure}[t]
    \centering
    \includegraphics[width=0.9\linewidth]{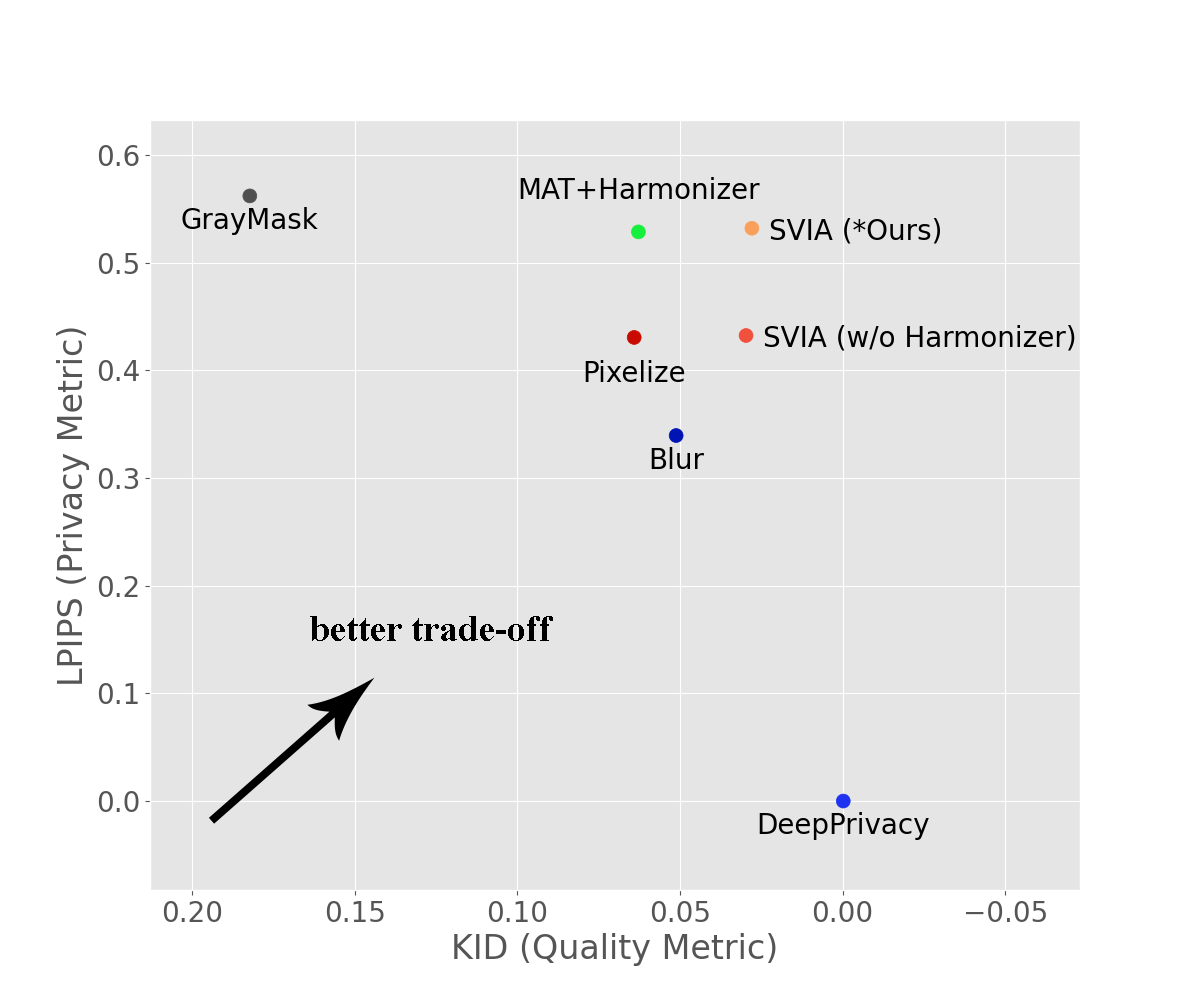}
    \caption{Illustration of the trade-off between image quality (KID) and privacy protection (LPIPS) for each method. The closer the point w.r.t.~a method is to the upper right corner, the more favorable its overall performance is.}
    \label{fig4}
\end{figure}

Fig.~\ref{fig4} shows a scatter plot that illustrates the overall performance of each method in terms of image quality and privacy protection.
The result signifies that SVIA best balances the quality of the generated images and the ability to prevent privacy leaks.
As evidenced by low FID and KID scores, high LPIPS scores, and low PerSim and ACR scores, SVIA is suitable for anonymizing street view images.

\begin{figure*}[t]
    \centering
    \includegraphics[width=0.875\linewidth]{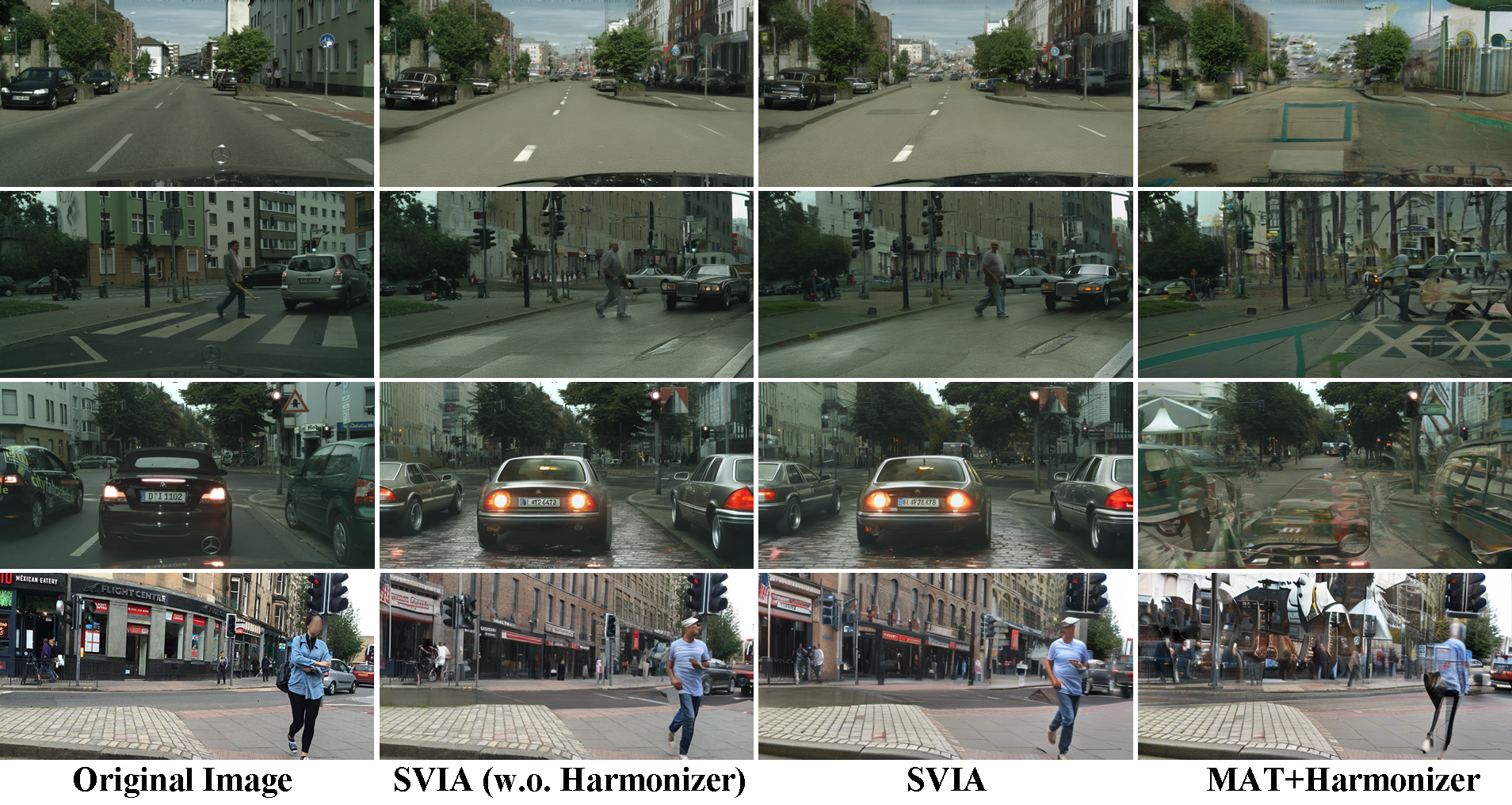}
    \caption{Visual comparison of four original images and their corresponding images generated by SVIA w/o Harmonizer, SVIA, and MAT+Harmonizer.}
    \label{fig5}
\end{figure*}
\begin{figure*}[t]
    \centering
    \includegraphics[width=0.875\linewidth]{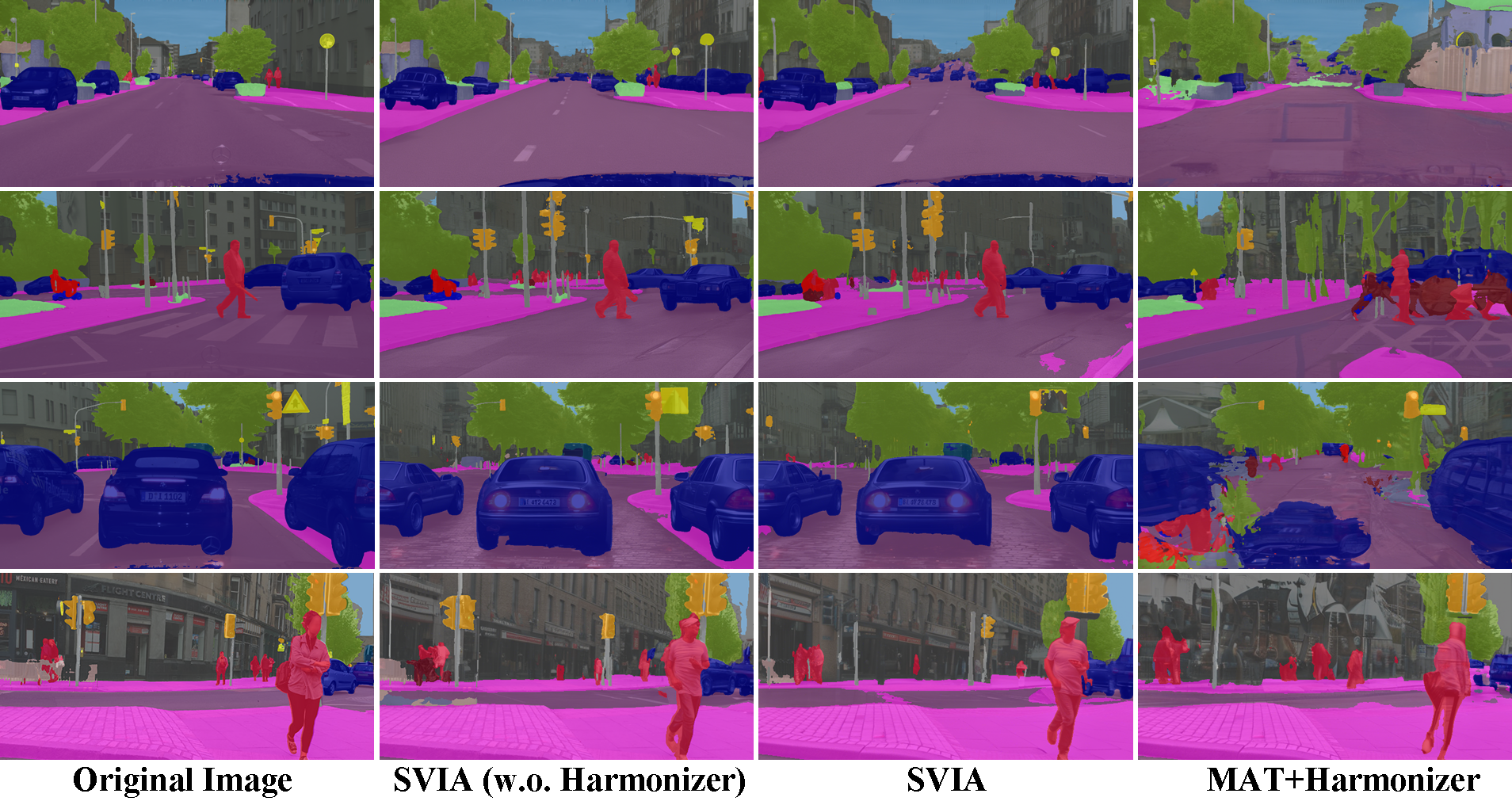}
    \caption{Illustration of the results for semantic segmentation on the images shown in Fig.~\ref{fig5}.}
    \label{fig6}
\end{figure*}

\begin{table}[t]
    \caption{Impact of the harmonizer on the performance of SVIA}
    \label{tbl-3}
    \begin{center}
    \scalebox{0.69}{
    \begin{tabular}{c|c|c|c|c|c}
        \hline
        \multirow{2}*{Dataset} & \multirow{2}*{Method} & \multicolumn{2}{c|}{Image Quality} & \multicolumn{2}{c}{Privacy Protection}\\
        \cline{3-6}
        ~ & ~ & FID $\downarrow$ & KID $\downarrow$ & LPIPS $\uparrow$ & PerSim $\downarrow$\\
        \hline
        \multirow{3}*{Cityscapes} & GrayMask & 163.1296 (3) & 0.1823 (3) & {\bf 0.5620 (1)} & \cellcolor[rgb]{0.85, 0.85, 0.85}{0.8845 (2)}\\
        ~ & SVIA w/o Harmonizer & \cellcolor[rgb]{0.85, 0.85, 0.85}{39.8516 (2)} & \cellcolor[rgb]{0.85, 0.85, 0.85}{0.0297 (2)} & 0.4323 (3) & 0.8861 (3)\\
        ~ & SVIA & {\bf 37.7365 (1)} & {\bf 0.0279 (1)} & \cellcolor[rgb]{0.85, 0.85, 0.85}{0.5320 (2)} & {\bf 0.8769 (1)}\\
        \hline
        \multirow{3}*{Mapillary Vistas} & GrayMask & 119.8208 (3) & 0.0668 (3) & \cellcolor[rgb]{0.85, 0.85, 0.85}{0.3575 (2)} & 0.8771 (3)\\
        ~ & SVIA w/o Harmonizer & {\bf 55.0217 (1)} & {\bf 0.0053 (1)} & 0.2795 (3) & \cellcolor[rgb]{0.85, 0.85, 0.85}{0.8727 (2)}\\
        ~ & SVIA & \cellcolor[rgb]{0.85, 0.85, 0.85}{65.0094 (2)} & \cellcolor[rgb]{0.85, 0.85, 0.85}{0.0105 (2)} & {\bf 0.4556 (1)} & {\bf 0.8595 (1)}\\
        \hline
    \end{tabular}
    }
    \begin{tablenotes}
        \item[1] Note: The meanings of bold font, gray boxes, and numbers in parentheses are the same as those of Table~\ref{tbl-1}.
    \end{tablenotes}
    \end{center}
\end{table}

\subsection{Ablation Study}

We conducted an ablation study to assess the significance of the harmonizer in SVIA.
We compare the generated images by SVIA with those directly output by the inpainter model using four quality and privacy measures.
Table~\ref{tbl-3} reports the results on the Cityscapes and Mapillary Vistas datasets.
The results reveal that the harmonizer improves image quality on the Cityscapes dataset but slightly degrades image quality on the Mapillary Vistas dataset.
Nevertheless, the harmonizer can further erase sensitive information and lead to better privacy protection on both datasets.

\subsection{Visual Results}
Fig.~\ref{fig5} presents four original images selected from the Cityscapes and Mapillary Vista datasets and their corresponding images generated by SVIA w/o Harmonizer, SVIA, and MAT+Harmonizer.
We can see that the images generated by SVIA are visually the closest to the original ones.
When enlarging the images generated without a harmonizer, we observe some ``hard lines'' between different semantic regions.
MAT can lead to obvious distortions in the generated images.
Fig.~\ref{fig6} shows the results for the semantic segmentation of the original and generated images.
We find that SVIA only slightly changes the segmentation results but MAT greatly degrades the segmentation quality.

\subsection{Time Efficiency}
It takes 7 and 30 days to run SVIA inference on the Cityscapes and Mapillary Vista datasets, respectively, in the default setting.
Then, the generation procedure to anonymize each image using SVIA takes about 2 minutes on both datasets.

\section{Conclusion}

In this paper, we propose SVIA, a novel framework to anonymize street view images collected for self-driving applications.
Our experimental results on the Cityscapes and Mapillary Vistas datasets demonstrate that SVIA achieves much better trade-offs between the quality of generated images and the level of privacy protection than several model-free and deep learning-based image anonymization methods.
In addition, SVIA has also been shown to have a less significant impact on downstream tasks than baselines.
Generally, SVIA has the potential to be deployed in self-driving applications because of its strong ability to protect privacy while not compromising the utility of images.

However, we note that SVIA still has some limitations.
First, SVIA cannot support video anonymization. Recently, the Stable Video Diffusion \cite{svd} and Sora \cite{sora} models enable the generation of high-resolution videos. They open up new possibilities for extending SVIA into the realm of video anonymization.
In the future, we would like to explore the ability of our pipeline to anonymize street-view videos.
Second, the generation speed of SVIA is relatively slow, which limits its application in real-time scenarios.
This issue might be mitigated by introducing the Stable Diffusion XL Turbo model \cite{sdxlturbo}.
Third, the architecture of SVIA can also be simplified by merging the inpainter with the harmonizer with the guidance of a semantic map \cite{ssmg}.


\bibliographystyle{IEEEtran}
\bibliography{refs}

\end{document}